\def\year{2019}\relax
\documentclass[letterpaper]{article} 
\usepackage{aaai19}  
\usepackage{times}  
\usepackage{helvet} 
\usepackage{courier}  
\usepackage[hyphens]{url}  
\usepackage{graphicx} 
\usepackage{graphicx}  
\usepackage{amssymb}
\usepackage{amsmath}

\title{Structured Two-stream Attention Network for Video Question Answering}
\author{Lianli Gao,\textsuperscript{\rm 1}
Pengpeng Zeng,\textsuperscript{\rm 1}
Jingkuan Song,\textsuperscript{\rm 1}
Yuan-Fang Li,\textsuperscript{\rm 2}
{\bf \Large Wu Liu,\textsuperscript{\rm 3}
Tao Mei,\textsuperscript{\rm 3}
Heng Tao Shen\textsuperscript{\rm 1}\thanks{Heng Tao Shen and Jingkuan Song are corresponding authors.}
}
\\
\textsuperscript{1}Center for Future Media and School of Computer Science and Engineering,\\
University of Electronic Science and Technology of China
\textsuperscript{2}Monash University
\textsuperscript{3}JD AI Research\\
\ lianli.gao@uestc.edu.cn,
\{is.pengpengzeng,jingkuan.song\}@gmail.com,
\{liuwu,tmei\}@live.com,
shenhengtao@hotmail.com}

\begin{document}
\maketitle
\begin{abstract}

To date, visual question answering (VQA) (i.e., image QA and video QA) is still a holy grail in vision and language understanding, especially for video QA. 
Compared with image QA that focuses primarily on understanding the associations between image region-level details and corresponding questions, video QA requires a model to jointly reason across both spatial and long-range temporal structures of a video as well as text to provide an accurate answer. 
In this paper, we specifically tackle the problem of video QA by proposing a Structured Two-stream Attention network, namely STA, to answer a free-form or open-ended natural language question about the content of a given video. 
First, we infer rich long-range temporal structures in videos using our structured segment component and encode text features. Then, our structured two-stream attention component simultaneously localizes important visual instance, reduces the influence of background video and focuses on the relevant text.
Finally, the structured two-stream fusion component incorporates different segments of query and video aware context representation and infers the answers. Experiments on the large-scale video QA dataset \textit{TGIF-QA} show that our proposed method significantly surpasses the best counterpart (i.e., with one representation for the video input) by 13.0\%, 13.5\%, 11.0\% and 0.3 for Action, Trans., TrameQA and Count tasks. It also outperforms the best competitor (i.e., with two representations) on the Action, Trans., TrameQA tasks by 4.1\%, 4.7\%, and 5.1\%.

\end{abstract}

\section{Introduction}
Recently, tasks involving vision and language have attracted considerable interests. Those include captioning ~\cite{DBLP:conf/aaai/GuC0C18,DBLP:conf/aaai/ChenDZH18,DBLP:conf/ijcai/SongGGLZS17,DBLP:journals/corr/abs-1708-02478} and visual question answering~\cite{DBLP:conf/iccv/AntolALMBZP15,DBLP:journals/corr/GaoMZHWX15,DBLP:journals/corr/RenKZ15,DBLP:conf/ijcai/SongZGS18,DBLP:conf/mm/GaoZSLS18}.
The task of captioning is to generate natural language descriptions of an image or a video.
On the other hand, visual question answering (VQA) (i.e., image QA and video QA) aims to provide the correct answer to a question regard to a given image/video. It has been regarded as an important Turing test to evaluate the intelligence of a machine \cite{lu2018rvqa}. The VQA problem plays a significant role in various applications, including human-machine interaction and tourist assistance. However, it is a challenging task, as it is required to understand both language and vision content to consider necessary commonsense and semantic knowledge, and to finally make reasoning to obtain the correct answer.

Image QA, which aims to correctly answer a question about an image, has achieved great progress recently.
Most existing methods for image QA use the attention mechanism \cite{DBLP:conf/iccv/AntolALMBZP15}, and they can be divided into two main types: visual attention and question attention.
The former attention focuses on the most relevant regions to correctly answer a question by exploring their relationships, which addresses ``where to look''.
The latter attention attends to specific words in the question about visual information, which addresses ``what words to listen to''.
Some works jointly perform visual attention and question attention \cite{DBLP:conf/nips/LuYBP16}.

In comparison, video QA is more challenging than image QA, as videos contain both appearance and motion information. The main challenges to video QA are threefold: first, a method needs to consider long-range temporal structures without missing important information; second, the influence of video background needs to be minimized to localize the correspond video instances; third, segmented information and text information need to be well fused. Therefore, we need more sophisticated video understanding techniques that can understand frame-level visual information and the temporal coherence during the progression of the video.
Video QA models also requires reasoning ability on spatial and long-range temporal structures of both video and text to infer an accurate answer.

Attention mechanisms has also been adopted for video QA, including spatial-temporal attention~\cite{DBLP:conf/cvpr/JangSYKK17} and co-memory attention~\cite{DBLP:journals/corr/abs-1803-10906}. 
Temporal attention learns which frames in a video to attend to, which is captured as \emph{whole-video} features. Co-memory proposes a co-memory attention mechanism: an appearance attention model to extract useful information from spatial features, and a motion attention model to extract useful cues from optical flow features. It concatenates the attended spatial and temporal features to predict the final results.

We observe that answering some questions in video QA requires focusing on many frames, which are equally important (e.g., How many times does the man step?).
Using only current attention mechanisms, and hence whole-video-level features, may ignore important frame-level information. Motivated by this observation, we introduce a new structure, namely structured segment, that divides video feature into $N$ segments and then takes each segment as input for a shared attention model.
Thus, we can obtain many important frames from multiple segments.
For better linking and fusing information from both video segments and the question, we propose a Structured Two-stream Attention network (STA) to learn high-level representations. Specifically, our model has two levels of decoders, where the first-pass decoder infers rich long-range temporal structures with our structured segment, and the second-pass encoder simultaneously localizes action instance and avoids the influence of background video with the assistance of structured two-stream attention.

Our STA model achieves state-of-the-art performance on a large-scale dataset: \emph{TGIF-QA} dataset.
To summarize, our major contributions include:
1) We propose a new architecture, Structured Two-stream Attention network (STA), for the video QA task by jointly attending to both spatial and long-range temporal information of a video as well as text to provide an accurate answer.
2) The rich long-range temporal structures in videos are captured by our structured segment component, while our structured two-stream attention component can simultaneously localize action instance and avoid the influence of background video.
3) Experimental results show that our proposed method significantly outperforms the state-of-the-arts in the Action, Trans. and FrameQA tasks on the \emph{TGIF-QA}. Notably, we represent our videos using only one type of visual features.

\section{Related Work}
\subsection{Image Question Answering}
Image QA~\cite{DBLP:conf/iccv/AntolALMBZP15,DBLP:journals/corr/GaoMZHWX15,DBLP:journals/corr/RenKZ15,DBLP:conf/aaai/LuLZWW18,DBLP:conf/cvpr/NamHK17,DBLP:conf/cvpr/YangHGDS16,DBLP:conf/eccv/XuS16,DBLP:conf/nips/LuYBP16,Patro_2018_CVPR,Teney_2018_CVPR}, the task that infers answers to questions on a given image, has achieved much progress recently.
Based on the framework of image captioning, most early works adopt typical CNN-RNN models, which use Convolutional Neural Networks (CNN) to extract image features and use Recurrent Neural Networks (RNN) to represent question information.
They integrate image features with question features using some simple fusion methods such as concatenation, summation and element-wise multiplication.
Finally, the fused features are fed into a softmax classifier to infer a correct answer. It has been observed that many questions are only related to some specific regions of an image, and various attention mechanisms has been introduced into image QA instead of using the global image features.
There are two main types of attention mechanisms: visual attention and question attention.
Specifically, visual attention learns which specific regions in the image to focus on for the question, while question attention attends to specific words in the question about vision information.
The work in \cite{DBLP:conf/cvpr/YangHGDS16} designs a Stacked Attention Networks which can search question-related image regions by performing multi-step visual attention operations.
In \cite{DBLP:conf/nips/LuYBP16,DBLP:conf/cvpr/NamHK17}, they present a co-attention mechanism that jointly performs question-guided visual attention and image-guided question attention to address the `which regions to look' and `what words to listen to' problems respectively. The typically used, simple fusion methods (e.g., concatenation, summation and element-wise multiplication) on visual and textual features cannot sufficiently exploit the relationship between images and questions.
To tackle this problem, some researchers introduced more sophisticated fusion strategies.
Bilinear (pooling) method \cite{DBLP:conf/cvpr/GaoBZD16} is one of the pioneering works to efficiently and expressively combine multimodal features by using an outer product of two vectors. Based on MCB \cite{DBLP:conf/cvpr/GaoBZD16}, lots of variants have been proposed, including MLB \cite{DBLP:journals/corr/KimOKHZ16} and MFB \cite{DBLP:conf/iccv/YuY0T17}.
The work in \cite{Nguyen_2018_CVPR} proposes a dense co-attention network (DCN) that computes an affinity matrix to obtain more fine-grained interactions between an image and a question.

\subsection{Video Question Answering}
Compared with image QA, video QA is more challenging.
The LSMDC-QA dataset \cite{DBLP:journals/ijcv/RohrbachTRTPLCS17} introduced the movie fill-in-the-blank task by transforming the LSMDC movie description dataset to the video QA domain.
The MovieQA dataset \cite{DBLP:conf/cvpr/TapaswiZSTUF16} aims to evaluate automatic story comprehension from both videos and movie scripts.
The work in \cite{DBLP:conf/cvpr/JangSYKK17} introduces a new large-scale dataset named \emph{TGIF-QA} and designs three new tasks specifically for video QA.
The attention mechanism has also been widely used in video QA.
The work in \cite{DBLP:conf/cvpr/YuKCK17} proposes a semantic attention mechanism, which detects concepts from the video first and then fuses them with text encoding/decoding to infer an answer.
The work in \cite{DBLP:conf/cvpr/JangSYKK17} proposes a dual-LSTM based approach with both spatial and temporal attention. 
The spatial attention mechanism uses the text to focus attention over specific regions in an image.
The temporal attention mechanism guides which frames in the video to look at for answering the question.
The work in \cite{DBLP:journals/corr/abs-1803-10906} proposes a co-memory attention network for video QA, which jointly models motion and appearance information to generate attention on both domains.
They introduce a method called dynamic fact ensemble to dynamically produce temporal facts in each cycle of fact encoding.
These methods usually extract compact whole-video-level features, while not adequately preserving frame-level information. However, a question to a video might be relevant to a sequence of frames, e.g., `How many times does the man step'. The compact whole-video-level features are not as informative as more fine-grained frame-level features. 
To address this issue, we propose to split a video into multiple segments in order to achieve a better balance between information compactness and completeness. 
We introduce our method in the next section.

\begin{figure*}[t]
  \centering\includegraphics[width=1.4\columnwidth]{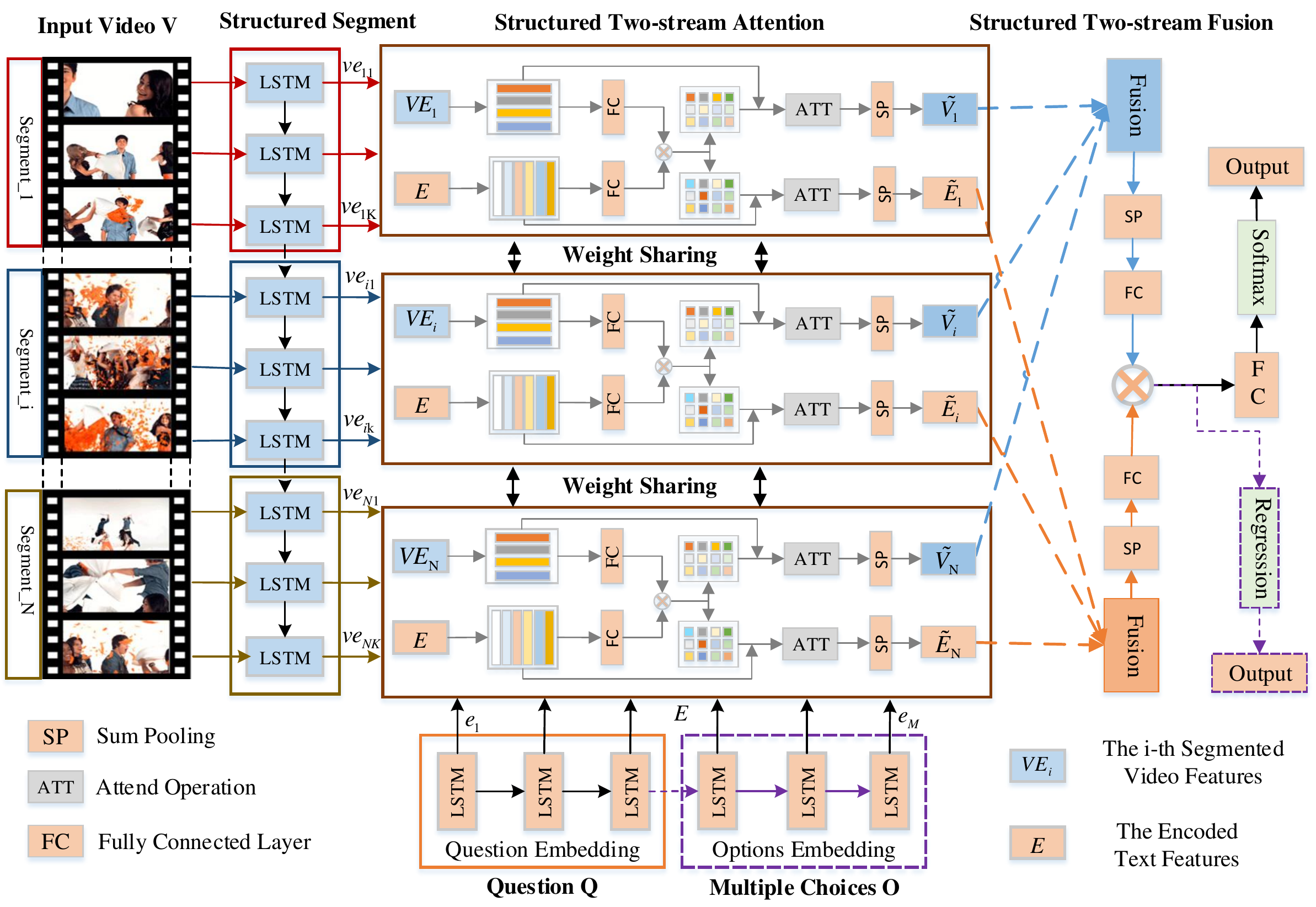}
  \caption{The framework of our proposed Structured Two-stream Attention Network (STA) for video QA. }\label{fig:framework}
\end{figure*}

\section{Methodology}
Recall that our aim is to efficiently extract video spatial and long-range temporal structures and then improve fusion of video and text representations to provide an accurate answer. As discussed in the Introduction, the primary challenges are threefold: (1) the incorporation of long-range temporal structures without missing important information; (2) the minimization of the influence of video background to localize the correspond video instances; and (3) the adequate fusion of segmented information with text information.

Our proposed framework is shown in Fig.\ref{fig:framework}. Formally, the input is a video $V$, a question $Q$ and a set of answer options $O$. In addition, only multiple choice type questions require the input of $O$, as shown in the purple dashed box in Fig.\ref{fig:framework}. Specifically, our framework consists of a number of \textit{a structured segments} that focuses on obtaining video long-range temporal information, \textit{a structured two-stream attention} that fuses language and video visual features repeatedly, on top of which is \textit{a structured two-stream fusion based answer prediction module} that fuses multi-modal segmental representations to predict answers. Below, we present the details of the above three major components.

\subsection{Structured Segment}
\textbf{Video Feature Extraction.}
Following previous work \cite{DBLP:journals/corr/abs-1803-10906,DBLP:conf/cvpr/YuKCK17}, we employ Resnet-152 \cite{DBLP:conf/cvpr/HeZRS16}, pre-trained on the ImageNet 2012 classification dataset \cite{DBLP:journals/corr/RussakovskyDSKSMHKKBBF14}, to extract video frame appearance features. More feature pre-processing details are given in Experiments section. For each video frame,  we obtain a $2048$-D feature vector, which is extracted from the pool5 layer and represents the global information of that frame. Therefore, an input video can be represented as:
\begin{eqnarray}
V = [v_{1}, v_{2}, ..., v_{T}], v_{t} \in  \mathbb{R}^{2048}
\end{eqnarray}
where $T$ is the length of a video.

For encoding sequential data streams, Recurrent Neural Networks (RNN) are widely and successfully used, especially in machine translation research. In this paper, we employ Long Short-term Memory (LSTM) networks to further encode video features  $\left\{ {{{\rm{v}}_{\rm{t}}}} \right\}_{{\rm{t = }}1}^{\rm{T}}$ to extract useful cues. For each step, e.g., \emph{t}-th step, the LSTM unit takes the $t$-th frame features and the previous hidden state ${\rm{h}}_{\rm{t-1}}^{\rm{v}}$ as inputs to output the $t$-th hidden state ${\rm{h}}_{\rm{t}}^{\rm{v}} \in {\mathbb{R}^D}$, where we set the dimension \emph{D} = 512.
\begin{eqnarray}
{\rm{h}}_{\rm{t}}^{\rm{v}}{\rm{ = LSTM(}}{{\rm{v}}_t}{\rm{,h}}_{t - 1}^v{\rm{)}}
\end{eqnarray}

\textbf{Structured Segment.} Previous work such as TGIF-QA \cite{DBLP:conf/cvpr/JangSYKK17} adopts a dual-layer LSTM to encode video features and then concatenates the last two hidden states of the dual-layer LSTM to represent whole-video-level information. This poses a risk of missing important frame-level information. To solve this problem, we introduce a new structure, namely structured segment, that firstly utilizes one-layer LSTMs to obtain $T$ hidden states ($\left\{ {{\rm{h}}_{\rm{t}}^v} \right\}_{{\rm{t = }}1}^{\rm{T}}$) and then divides the $T$ hidden states into $N$ segments ($\{ V{E_i}\} _{i = 1}^N$). After this stage, a video can be represented as $\{ V{E_i}\} _{i = 1}^N$, and the $i$-th segment  ${VE}_i$  is represented as: 
 \begin{eqnarray}
 {VE}_i = \{ v{e_{i1}},v{e_{i2}},...,v{e_{iK}}\}, v{e_{ik}} \in  \mathbb{R}^{512}
 \end{eqnarray}
where $v{e_{ik}}$ is the hidden states of the $k$-th frame in the $i$-th segment, $K$ the total number of hidden states for each segment. The value of $K$ is the same for each segment.

\subsection{Text Encoder}
For our video QA task, there are two types of questions: open-ended question and multiple-choice question. For the first type, our framework takes only the question as the text input, while for the second type, our framework takes both the question and answer options as the text input. The final text feature is represented as $E$.

\textbf{Question Encoding.}
A question, consisting of $M$ words, is first converted into a sequence $Q=\left\{ {{q_m}} \right\}_{m = 1}^M$, where $q_{m}$ is a one-hot vector representing the word at position $m$. Next, we employ the word embedding GloVe \cite{DBLP:conf/emnlp/PenningtonSM14} pre-trained on the Common Crawl dataset, to process each word to obtain a fixed word vector. After GloVe, the $m$-th word is represented as $x_m^q$. Next, we utilize a one-layer LSTM on top of the word embeddings to model the temporal interactions between words. The LSTM takes the embedding vectors $\left\{{x_m^q}\right\}_{m = 1}^M$ as inputs, and finally we obtain a question feature $E$ for answer prediction process. As a result, the question encoding process can be defined as below:
\begin{eqnarray}
x_m^q &=& W_e^q{q_m}\\
e_m^q &=& LSTM(x_{\rm{m}}^{\rm{q}},e_{m - 1}^q)
\end{eqnarray}
where $W_e^q$ is an embedding matrix. The dimension of all the LSTM hidden states is set to \emph{D} = 512. Finally, after question encoder, $Q$ is represented as $E = \left\{ {{e_1^q}, \cdots ,{e_M^q}} \right\}$.

\textbf{Multi-choice Encoding.}
For the task of Multi-choice, the input involves a question and  a set of answer candidates. To process answer candidates, we follow the above question encoding procedure to transform each word of the options into a one-hot vector and then further embed it the GloVe. We consider answer candidates as complementary to the question. Therefore, the text input of our framework becomes $Q' = \left[ {Q,O} \right]$, where $O$ is the answer candidate features and  $[,]$ represents concatenation. Furthermore, the one-layer LSTM unit takes the merged $Q'$ as the input to extract text feature $E$. We formulate this encoding process as below:
\begin{align}
Q' &= \left[ {Q,O} \right] = \left\{ {{q'_1}, \cdots ,{q'_{M}}} \right\}\\
x_{m}^q &= W_e^q{q'_{m}}\\
e_{m}^q &= LSTM\left( {x_{m}^q,e_{{m} - 1}^q} \right)
\end{align}
where $M$ is the sum of the length of question words and the length of all candidate words. Finally, after the multi-choice encoder, $Q'$ is represented as $E = \left\{ {{e_{1}^q}, \cdots ,{e_{M}^q}} \right\}$. 

\subsection{Structured Two-stream Attention Module}
We now describe the second major component, the Structured Two-stream Attention layer (seen Fig.\ref{fig:framework}), which links and fuses information from both video segments and text. This attention layer consists of $N$ two-stream (i.e., text and video features) attentions and all the attention models share parameters. 

For the $i$-th two-stream attention model, it takes the $i$-th segmented video encoded feature $V{E_i}$ and text feature $E$ as input to learn interactions between them to update both $V{E_i}$ and $E$. Here, we denote the input to the $i$-th two-stream attention by $V{E_i} = \left\{ {v{e_{i1}}, \cdots ,v{e_{iK}}} \right\} \in {\mathbb{R}^{D \times K}}$ and $E = \left\{ {{e_1}, \cdots ,{e_M}} \right\} \in {\mathbb{R}^{D \times M}}$. Unlike previous video QA methods such as TIGF-QA \cite{DBLP:journals/corr/abs-1803-10906} and Co-memory \cite{DBLP:journals/corr/abs-1803-10906}, which simply concatenate video frame features with text question features to form a new feature for answer prediction, our two-stream attention mechanism calculates attention in two directions: from video to question as well as from question to video. Both attention scores are computed from a shared affinity matrix ${A_i}$, which is computed by:
\begin{eqnarray}
\label{eqn}
{A_i} = {\left( {V{E_i}} \right)^T}{W_s}{E} 
\end{eqnarray}
where ${W_s}$ is a learnable weight matrix. For the convenience
of calculation, we replace Eq.(\ref{eqn}) by two separate linear projections. Thus, Eq.(\ref{eqn}) is re-defined as below:
\begin{eqnarray}
{A_i} = {\left( {{W_v}V{E_i}} \right)^T}\left( {{W_q}{E}} \right)
\end{eqnarray}
where $W_v$ and $W_q$ are linear function parameters. In essence, ${A_i}$ encodes the similarity between its two inputs $V{E_i}$ and $E$. With ${A_i}$, we can compute attentions and then attend to the two-stream features in both directions.

\textbf{1st-stream: Visual Attention.} Visual attention vector indicates \textit{which frames in a video shot to attend to or most relevant to each question word}. Given ${A_i} \in {\mathbb{R}^{K \times M}}$, the attention vector is computed by the following function:
\begin{eqnarray}
{C_i} = soft\max ({{\max}_{col}}{\left( {{A_i}} \right)^T})
\end{eqnarray}
where ${C_i} \in {\mathbb{R}^{K \times 1}}$, $\max _{col}$ indicates column-wise max operation on ${A_i}$. After column-wise max operation, we use \textit{softmax} operation to normalize the value to produce the attention vector ${C_i}$. Specifically, $\sum\nolimits_{k = 1}^K {{c_{ik}}}  = 1$. Next, we conduct the following operation to obtain the attend video feature ${{\tilde V}_i}$:
\begin{eqnarray}
{{\tilde V}_i} = \sum\nolimits_{k=1}^K {{c_{ik}}v{e_{ik}}}
\end{eqnarray}
where ${{\tilde V}_i} \in {\mathbb{R}^{1 \times D}}$, which contains the attended visual vectors respect to the entire question. 

\textbf{2nd-stream: Text Attention.} Textual attention vector indicates \textit{which question word to attend to}. Given ${A_i} \in {\mathbb{R}^{K \times M}}$, we normalize ${A_i}$ in row-wise with \textit{softmax} to derive the attention map on question words conditioned by each video frame. Formally, the attention vector is computed by the following function:
\begin{eqnarray}
{B_i} = softmax \left( {{A_i}} \right)
\end{eqnarray}
where ${B_i} \in {\mathbb{R}^{K\times M}}$. With the generated attention map ${B_i}$, we utilize it to attend to the question words to produce a more representative question feature ${{\tilde E}_i}$. Formally, ${{\tilde E}_i}$ is computed by the following function:
\begin{eqnarray}
{{\tilde E}_i} = {B_i}{E^T}
\end{eqnarray}
where ${{\tilde E}_i} \in {\mathbb{R}^{K\times D}}$. Finally, we conduct column-wise sum to obtain the final ${{\tilde E}_i} \in {\mathbb{R}^{1\times D}}$. Hence the generated ${\tilde E}$ contains the attended question words vectors for the entire video segments.

\subsection{Structured Two-stream Fusion}
After computing the attended feature representations ${\tilde V}$ and ${\tilde E}$, where $\tilde V = \left\{ {{{\tilde V}_1}, \cdots ,{{\tilde V}_N}} \right\}$ and $\tilde E = \left\{ {{{\tilde E}_1}, \cdots ,{{\tilde E}_N}} \right\}$, we first fuse them respectively.
\begin{eqnarray}
\tilde V &=& \sum\nolimits_{i = 1}^N {{{{\rm{\tilde V}}}_i}} \\
\tilde E &=& \sum\nolimits_{i = 1}^N {{{\tilde E}_i}} 
\end{eqnarray}
Followed by a sum pooling, the attended video vector ${\tilde V}$ and attended question vector ${\tilde E}$ are pooled and combined together to yield $H$ for answer prediction.
\begin{eqnarray}
H{\rm{ }} = Relu({W_{fv}}\tilde V + {b_v}) \otimes Relu({W_{fq}}\tilde E + {b_q})
\end{eqnarray}
where $W_{fv}$ and $W_{fq}$ are parameters; $b_v$ and $b_q$ are bias terms; $\otimes$ is the element-wise multiplication, $Relu$ is the activation function. 

\subsection{Answer Decoder Moduler}
Our output layer (i.e.\ answer decoder) is application-specific. Our framework allows us to easily swap the output layer based on the task type with the rest of the architecture remaining exactly the same. Following previous work \cite{DBLP:conf/cvpr/JangSYKK17,DBLP:journals/corr/abs-1803-10906}, we treat the four tasks (i.e.\ Count, Action, State Trans., and FrameQA) in the \emph{TGIF-QA} dataset as three different type decoders: multiple choice, open-ended numbers and open-ended words. Here, we describe the three types of answer decoders for each specific video QA task. 

\textbf{Multiple Choice.} For the TGIF-QA dataset, both State Trans.\ and Action tasks belong to the multiple choice category  QA. In order to solve both tasks, we apply a linear regression function for the above final output $H$ and derive a real-valued score for each answer option.
\begin{eqnarray}
s = {W_h}^\prime H
\end{eqnarray}
where $W_h^\prime$ are weight parameters. We define $s_p$ and $s_n$ as the real-valued scores derived from the positive and negative answers, respectively. In order to train our mode, we minimize the hinge loss of pair comparisons, $\max (0,1 + {s_n} - {s_p})$.

\textbf{Open-ended Numbers.} For the TGIF-QA dataset, the Count task requires a model to count the number of repetitions of an action, and the answer ranges from $0$ to $10$. For this task, we define a linear regression function which takes a predicted output $H$ as input and produces an integer-valued answer, ranging from $0$ to $10$. The output score is defined as:
\begin{eqnarray}
s = [{{W'}_h}H + b_h]
\end{eqnarray}
where $[.]$ means rounding, ${W'}_h$ are model parameters and $b_h$ is the bias. In order to train the model, we adopt the Mean Square Error (MSE) loss between the real value and the predicted value.

\textbf{Open-ended Words.} Similar to the task of image QA, we treat video FrameQA as a multi-class classification problem, in which each class corresponds to a distinct answer (i.e., a dictionary word, a type of object, color, number or location). For video QA, we use a softmax classifier to infer the final answer. The candidate with the highest probability is considered as the final answer. This output is defined as:
\begin{eqnarray}
  o = softmax({{W'}_h}H + b_h)
\end{eqnarray}
where ${W'}_h$ are weight parameters and $b_h$ is the bias. The model is optimized by minimizing the cross-entropy loss.

Note that, in this paper we deal with four tasks. For each task, we separately train each model with the corresponding answer decoder and loss function mentioned above. Eventually, each model is evaluated separately.

\section{Evaluation}
\label{sec.exp}
In this section, we first describe the dataset, evaluation metrics and implementation details. Then we report and analyze the experimental results. 

\begin{table}[]
	\small
  \centering
  \caption{Statistics of the TGIF-QA dataset.}
  \label{tab1}
  \begin{tabular}{|c|c|c|c|}
    \hline
    Task             & Train   & Test   & Total   \\ \hline
    Repetition Count & 26,843  & 3,554  & 30,397  \\ \hline
    Repeating Action & 20,475  & 2,274  & 22,749  \\ \hline
    State Transition & 52,704  & 6,232  & 58,936  \\ \hline
    Frame QA         & 39,392  & 13,691 & 53,083  \\ \hline
    Total            & 139,414 & 25,751 & 165,165 \\ \hline
  \end{tabular}
\end{table}

\subsection{Dataset and Evaluation Settings}
To evaluate the performance of the video QA models, we follow two recent video QA work \cite{DBLP:conf/cvpr/JangSYKK17,DBLP:journals/corr/abs-1803-10906} to evaluate our method on the large-scale public video QA dataset TGIF-QA.

\textbf{TGIF-QA Dataset.} It is a large-scale dataset collected by \cite{DBLP:conf/cvpr/JangSYKK17}, which is designed specifically for video QA to better evaluate a model's capacity for deeper video understanding and reasoning. Jang \textit{et al.} collected 165k QA pairs from 71k animated Tumblr GIFs \cite{DBLP:conf/cvpr/LiSCTGJL16} and defined four task types: repetition count (Count), repeating action (Action), state transition (Trans.) and video frame QA (FrameQA). Each of them has approximately 30k, 22k, 58k and 53k questions, respectively. The specific setting (e.g., train and test splits) are demonstrated in Tab.\ref{tab1}, following the original setting of \cite{DBLP:conf/cvpr/JangSYKK17}. Compared with FrameQA which can be answered by analyzing the content of one particular video frame, other three tasks Count, Action and Trans.\ can only be answered accurately only via reasoning across multiple frames. In addition, both Count and FrameQA contain open-ended questions, while Action and Trans.\ contain multi-choice questions.

\textbf{Four Tasks Settings.} In this paper, we deal with four specific video QA tasks: Count, Action, Trans. and FrameQA. Specifically, each Count question has 11 possible answers, ranging from zero to ten. It requires a model to calculate how many times an action has been repeated. For instance, ``How many times does the woman chew food?''. Compared with Count, Action focuses on the actions appearing in a video. Instead of asking how many times an action is performed, it asks which action is repeated a given number of times in a video. For example, ``What does the woman do 4 times?''. For Trans., each question provides five candidate options and all the questions are about the understanding of the transition of two action states in a video. For instance, ``What does the woman do before places hands in lap?''. For FrameQA, it is similar to image QA, but more complex. To provide an accurate answer, a model must locate the specific frame and its specific regions that the question refers to. For example, ``what is the color of the hair?''. We treat FrameQA as a multi-class classification problem.

\textbf{Evaluation Metric.} Following \cite{DBLP:conf/cvpr/JangSYKK17,DBLP:journals/corr/abs-1803-10906}, we use the same evaluation metrics. Specifically, FrameQA, Action and Trans., the accuracy (ACC.) is employed to evaluate model performance. Thus, the higher the ACC.\ is, the better the model is. The performance of Count is measured by Mean Square Error (MSE) between the true answer and the predicted answer. Therefore, lower MSE value indicates better performance of the model.  
\begin{table}[]
	\small
  \centering
  \caption{Ablation study on the \emph{TGIF-QA} dataset. For evaluation metric, Action, Trans. and FrameQA use ACC (\%), while Count adopts Mean Square Error (MSE). V and T indicates visual attention and text attention. $N$ indicates the number of video segments. ST is the best published methods using the same features as our model uses.}
  \label{tab2}
  \begin{tabular}{l||llll}
    \hline
    Method   & Action         & Trans          & Frame        & Count         \\ \hline
    STA-V-T(N=1) & 71.94          & 78.67          & 56.01          &   4.26        \\ 
    STA-V-T(N=2) & 72.16          & 78.84          & 55.92          &    4.25       \\ 
    STA-V-T(N=3) & \textbf{72.38} & 78.96          & 56.52          &     4.27      \\ 
    STA-V-T(N=4) & 72.34          & \textbf{79.03} & \textbf{56.64} &  \textbf{4.25}\\ \hline 
    
    STA-V(N=1)  & 71.02          & 77.24          & 54.89          & 4.32          \\ 
    STA-V(N=2)  & 71.46          & 77.47          & 55.05          & 4.37          \\ 
    STA-V(N=3)  & 71.68 & 77.31          & 55.50 & 4.34          \\ 
    STA-V(N=4)  & 71.24          & 77.57 & 55.47          & 4.33 \\ \hline
    ST \cite{DBLP:conf/cvpr/JangSYKK17}         & 59.04   & 65.56  & 45.60 & 4.55  \\ \hline
  \end{tabular}
\end{table}
\subsection{Implementation Details}
For fair comparisons with recent work \cite{DBLP:conf/cvpr/JangSYKK17,DBLP:journals/corr/abs-1803-10906}, we use the same network  ResNet152 \cite{DBLP:conf/cvpr/HeZRS16} pre-trained on ImageNet 2012 classification to extract frame features. More specifically, all the frame features are obtained from the same pooling layer (pool5) and their dimension is $2048$. In all tables in the experimental section, we use \textit{R} to indicates that the input video's feature is extracted from ResNet152 feature. In addition, to reduce redundant information and reduce computation cost, we sample 36 frames from each video with equal spacing.

For text representation, we first encode each word with a pre-trained GloVe embedding to generate a 300-D vector following \cite{DBLP:conf/cvpr/JangSYKK17,DBLP:journals/corr/abs-1803-10906}. All the words are further encoded by a one-layer LSTM, whose hidden state has the size of 512. All the hidden states are concatenated and used for co-attention.  

\textbf{Training Details.} In our experiments, the optimization algorithm is Adamax. The batch size is set as 128. The train epoch is set as 30. In addition, gradient clipping, weight normalization and dropout are employed in training. In addition, our implementation is based on the Pytorch library.

\begin{figure*}[t]
  \centering\includegraphics[width=1.5\columnwidth]{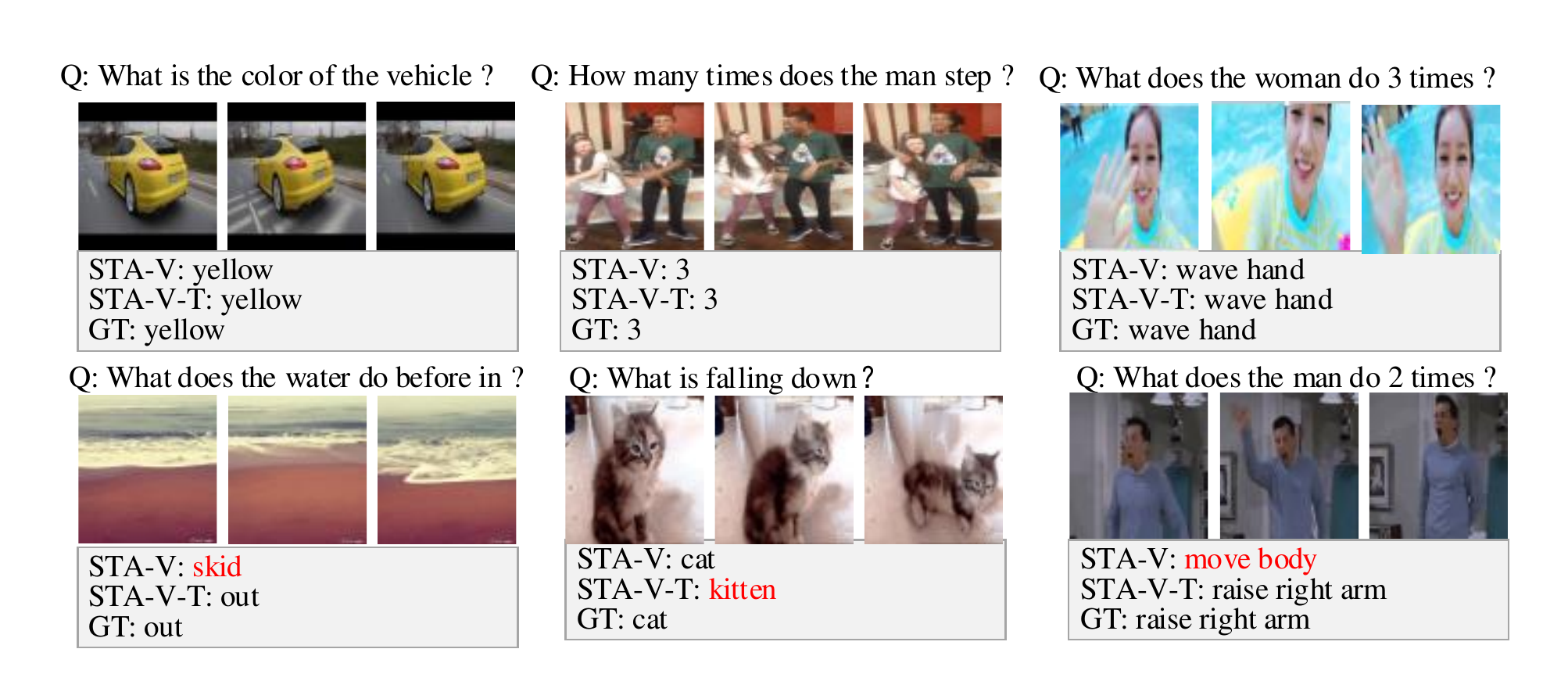}
  \caption{Some qualitative results produced by our models. Text in orange indicates incorrect answers.}\label{fig:vis}
\end{figure*}

\subsection{Ablation Study}
The framework of our proposed STA consists of multiple major components. In order to evaluate the contribution of each component to the final performance, we conduct several ablation studies on the \emph{TGIF-QA} dataset. Experimental results are shown in Tab.~\ref{tab2}. 

\textbf{The Role of Structured Segmentation.}
The first block of Tab.~\ref{tab2} shows the effect of $N$, which is the number of structured segments. From the first block, we found that $N=4$ improves performance of all four tasks to a certain extent. One possible reason is that dividing a video into multiple segments to conduct attention has the potential to locate the most relevant frames as well as to learn the long structures. However, the performance improvement with the increase in $N$ is minor and the reason might be the nature of the GIF videos, which are well segmented and carefully curated, with an average length of 47 frames. Thus the advantages of our structured segment are not fully exploited. 

\textbf{The Role of Two-stream Attention.}
To analyze the contribution of two-stream attention: 1st-stream visual attention and 2nd-stream text attention, we conduct the second ablation study by removing the text attention and keeping the visual attention, which is represented as STA-V in Tab.\ref{tab2}. From the table, we can see that with the same setting of $N$, STA-V-T performs better than STA-V. When $N=4$, STA-V-T surpasses STA-V by 1.1\% on Action, 1.46\% on Trans.\ and 1.17\% on FrameQA, and reduces MSE to 4.25 on Count. This ablation study shows the beneficial effects of our text attention. 

In order to examine the influence of the visual attention, we compare our STA-V  with the best published results obtained by the spatial-temporal reasoning (ST) method \cite{DBLP:conf/cvpr/JangSYKK17}. From the last block we can see that our STA-V significantly outperforms ST by a large margin (12.24\%, 12.07\%, 9.87\% on Action, Trans. and FrameQA, respectively). In addition, for Count, compared with ST with MSE of 4.55, STA-V-T reduces the error score to 4.33. 
\begin{table}[]
  \centering
  \small
  \caption{Comparison with the state-of-the-art method on the \emph{TGIF-QA} dataset. R indicates ResNet152 features. For video representation, all methods take video spatial vectors only as the visual inputs.}
  \label{tab3}
  \begin{tabular}{l|llll}
    \hline
    Model          & Action & Trans. & Frame & Count \\ \hline\hline
    VIS+LSTM(agg)(R) & 46.8   & 56.9  & 34.6  & 5.09  \\
    VIS+LSTM(avg)(R)& 48.8   & 34.8  & 35.0  & 4.80  \\ 
    VQA-MCB(agg)(R) & 58.9   & 34.3  & 25.7  & 5.17  \\
    VQA-MCB(avg)(R)& 29.1   & 33.0  & 15.5  & 5.54  \\ 
    CT-SAN(R)& 56.1   & 64.0  & 39.6  & 5.13  \\ 
    ST(R) & 59.0   & 65.5  & 45.6 & 4.55  \\ \hline
    \textbf{STA(R)}     & \textbf{72.3}  & \textbf{79.0} & \textbf{56.6} & \textbf{4.25}  \\ \hline
  \end{tabular}
\end{table}

\begin{table}[]
  \centering
  \small
  \caption{Comparison with the state-of-the-art multi-feature based methods on the \emph{TGIF-QA} dataset. R, C and F indicate Resnet152, C3D and Optical Flow features, respectively. Sp and Tp indicate spatial attention and temporal attention respectively.}
  \label{tab4}
  \begin{tabular}{l|llll}
    \hline
    Model          & Action & Trans & Frame & Count \\ \hline\hline
    ST (R+C)        & 60.1   & 65.7  & 48.2  & 4.38  \\
    ST-Sp (R+C)     & 57.3   & 63.7  & 45.5  & 4.28  \\
    ST-Sp-Tp (R+C)  & 57.0   & 59.6  & 47.8  & 4.56  \\
    ST-Tp (R+C)     & 60.8   & 67.1  & 49.3  & 4.40  \\
    ST-Tp (R+F)     & 62.9   & 69.4  & 49.5  & 4.32  \\ 
    
    Co-memory (R+F)      & 68.2   & 74.3  & 51.5  & \textbf{4.10}  \\ \hline
    \textbf{STA (R)}     & \textbf{72.3}  & \textbf{79.0} & \textbf{56.6} & 4.25 \\ \hline
    
  \end{tabular}
\end{table}

\subsection{Qualitative Results}
To understand the effect of our attention mechanism, we show some examples in Fig.~\ref{fig:vis}. The first row demonstrates three positive examples, where both models (i.e., STA-V and STA-V-T) can provide correct answers. The second row shows other examples. More specifically, the first and the third examples show that STA-V-T answers the questions correctly, while STA-V fails. The middle examples shows that occasionally STA-V can provide the correct answer.

\subsection{Comparing with State-of-the-Arts on the TGIF-QA Dataset}
To date, a few studies have been conducted on the video QA task. Existing methods can be divided into two categories: 1) image-based approaches including VIT+LSTM \cite{DBLP:conf/nips/RenKZ15} and VQA-MCB \cite{DBLP:conf/emnlp/FukuiPYRDR16}; and 2) video-based approaches including ST \cite{DBLP:conf/cvpr/JangSYKK17} and Co-memory \cite{DBLP:journals/corr/abs-1803-10906}. Our method belongs to the second category. Usually, videos contain two types of features: spatial and temporal. Spatial features are usually extracted from a CNN network, while temporal features are extracted from optical flows or via a C3D network. Therefore, in this case, we can divide the existing methods into two categories: single feature based methods and multi-feature based methods.

Our method STA utilizes single feature for representing videos. Therefore, in this section, we first compare our method with existing methods which take only ResNet152 frame features as visual inputs to conduct answer prediction. The experimental results are shown in Tab.~\ref{tab3}. In order to apply image-based approach, multiple frames' spatial features must be firstly merged into a vector, which is considered as the alternative of image features to be fed into an image-based agg.\ or avg.\ based model. Agg.\ and avg.\ are proposed in \cite{DBLP:conf/cvpr/JangSYKK17} to directly apply image-based methods. The agg.\ is conducted by averaging all frames' features and uses the average spatial feature as input to the model. Compared with agg., avg.\ is more complicated. It runs answer prediction on each frame, and then averages all frames' predicted answers to obtain the final result. It can be seen from Tab.~\ref{tab3} that our method outperforms the best publish results ST(R) by 13.3\%, 13.5\% and 11.0\% on Action, Trans.\ and FrameQA, respectively. For count, STA reduces the error score to 4.25.

Furthermore, we compare our method STA with multi-feature based methods, which include ST \cite{DBLP:conf/cvpr/JangSYKK17} and Co-memory \cite{DBLP:journals/corr/abs-1803-10906}. Specifically, ST combines ResNet152 features (marked as R in Tab.\ref{tab4}) with C3D features (C) or optical flow features (F), while Co-memory incorporates both R and F. The comparison results are given in Tab.~\ref{tab4}. From these results, we can observe that Co-memory (R+F) outperforms all variants of ST, and achieves the best performance on the Count task. 
Compared with Co-memory (R+F), our STA model only takes ResNet152 features as input. However, for Action, Trans.\ and FrameQA tasks, it significantly outperforms Co-memory (R+F) (68.2\%, 74.3\%, and 51.5\%, respectively) and yields the best performance reaching 72.3\%, 79.0\%, 56.6\%, respectively. Even for the Count task, our STA achieves the second best and comparable result.

\section{Conclusion}
In this work, we propose a novel Structured Two-stream Attention Network (STA) for Video Question Answering.
We first utilize our structured segment component to infer rich long-range temporal structures in videos and also encode questions to features.
Then, instead of simply concatenating video and question features to answer the questions, we utilize a two-stream attention mechanism to simultaneously localize visual instances relevant to the questions and to avoid the influence of background video or irrelevant text.
Finally, the structured two-stream fusion component incorporates different segments of query- and video-aware context representations and infers the answers for different types of questions.
Our comprehensive evaluation on the \emph{TGIF-QA} dataset shows our STA model outperforms state-of-the-art methods significantly.

\section*{Acknowledgements}
This work is supported by the Fundamental Research Funds for the Central Universities (Grant No. ZYGX2014J063, No. ZYGX2016J085), the National Natural Science Foundation of China (Grant No. 61772116, No. 61502080, No. 61632007, No. 61602049), and JD AI Research.

\bibliographystyle{aaai}
\bibliography{aaai2019}
\end{document}